# English Sentence Recognition using Artificial Neural Network through Mouse-based Gestures

Firoj Parwej
Research Scholar Ph. D (Computer Science)
Department of Computer Science & Engineering,
CMJ University, Shilong, Meghalaya, India

## ABSTRACT

Handwriting is one of the most important means of daily communication. Although the problem of handwriting recognition has been considered for more than 60 years there are still many open issues, especially in the task of unconstrained handwritten sentence recognition. This paper focuses on the automatic system that recognizes continuous English sentence through a mouse-based gestures in real-time based on Artificial Neural Network. The proposed Artificial Neural Network is trained using the traditional backpropagation algorithm for self supervised neural network which provides the system with great learning ability and thus has proven highly successful in training for feed-forward Artificial Neural Network. The designed algorithm is not only capable of translating discrete gesture moves, but also continuous gestures through the mouse. In this paper we are using the efficient neural network approach for recognizing English sentence drawn by mouse. This approach shows an efficient way of extracting the boundary of the English Sentence and specifies the area of the recognition English sentence where it has been drawn in an image and then used Artificial Neural Network to recognize the English sentence. The proposed approach English sentence recognition (ESR) system is designed and tested successfully. Experimental results show that the higher speed and accuracy were examined.

## 1. INTRODUCTION

The Handwriting recognition refers to the identification of written characters. Handwriting recognition has become an acute research area in recent years for the ease of access of computer applications. Numerous approaches have been proposed for character recognition and considerable successes have been reported [1]. Traditional handwritten character recognition techniques enable a computer to receive and interpret intelligible handwritten input from sources such as papers, documents, touch-screens or pictures. During the last years, many popular studies and applications merged for bank check processing, mailed envelops reading , and handwritten text recognition in documents and videos [2]. Until now, it is still a difficult task for a machine to recognize human handwritings with significant accuracy, especially under variable circumstances such as variations in writings, variable sizes, and different patterns for different people etc. The English comprises of 26 basic alphabets which are simple to write and recognize but becomes very complex when they are handwritten. In this paper we are using a supervised neural network to do the work of classification of basic alphabets and have trained it using Back-Propagation algorithm. The mouse-based interaction, we decided to simplify the definition, while still keeping the notion usable a mouse gesture is a continues, directed sequence of the mouse cursor movements with the clearly distinguished start and end [3]. In this paper First of all the image database was created consisting of 26 alphabets each dynamically hand drawn and stored after that English sentence to be recognized is drawn with a mouse. This image is then normalized, followed by feature extraction. The extracted features are then fed as input to the Artificial Neural Network for recognition. Artificial Neural network approaches have been used in many classification problems [4]. The traditional classifiers which tend to test competing hypotheses sequences, neural networks test the competing hypotheses in parallel, thus providing high computation rates.

## 2. ENGLISH WRITTEN CHARACTERISTICS

The English language belongs to the West Germanic branch of the Indo-European family of languages. The closest undoubted living relatives of English are Scots and Frisian. Frisian is a language spoken by approximately half a million people in the Dutch province of Friesland, in nearby areas of Germany, and on a few islands in the North Sea. The history of the English language has traditionally been divided into three main periods: Old English (450-1100 AD), Middle English (1100-circa 1500 AD) and Modern English (since 1500). Over the centuries, the English language has been influenced by a number of other languages.

Old English (450 - 1100 AD): During the 5th Century AD three Germanic tribes (Saxons, Angles, and Jutes) came to the British Isles from various parts of northwest Germany as well as Denmark. These tribes were warlike and pushed out most of the original, Celtic-speaking inhabitants of England into Scotland, Wales, and Cornwall. One group migrated to the Brittany Coast of France where their descendants still speak the Celtic Language of Breton today [5].

Middle English (1100-circa 1500 AD): After William the Conqueror, the Duke of Normandy, invaded and conquered England in 1066 AD with his armies and became king, he brought his nobles, who spoke French, to be the new government. The Old French took over as the language of the court, administration, and culture. Latin was mostly used for writing the language, especially that of the Church. Meanwhile, The English language, as the language of the now lower class, was considered a vulgar tongue.

Modern English (1500 to the present): Modern English developed after William Caxton established his printing press at Westminster Abbey in 1476. Johann Gutenberg invented the printing press in Germany around 1450, but Caxton set up

England's first press. The Bible and some valuable manuscripts were printed. The invention of the printing press made books available to more people. The books became cheaper and more people learned to read. Printing also brought standardization to English. Since around the 9th century, English has been written in the Latin script, which replaced Anglo-Saxon runes. The modern English alphabet contains 26 letters of the Latin script. Early Modern English and Late Modern English vary essentially in vocabulary. Late Modern English has many more words, arising from the Industrial Revolution and the technology that created a need for new words as well as international development of the language [6].

The Britain was an Empire for 200 years between the 18th and 20th centuries and English language continued to change as the British Empire moved across the world - to the USA, Australia, New Zealand, India, Asia and Africa. They sent people to settle and live in their conquered places and as settlers interacted with the natives, new words were added to the English vocabulary. The British Empire at its height covered one quarter of the Earth's surface, and the English language adopted foreign words from many countries. British English and American English, the two major varieties of the language, are spoken by 400 million persons. Received Pronunciation of British English is the prestige variety, while General American English is more influential. The total number of English speakers worldwide may exceed one billion.

## 3. RESEARCH METHODOLOGY

Artificial neural networks (ANN) have been developed as generalizations of mathematical models of biological nervous systems. A first wave of interest in neural networks (also known as connectionist models or parallel distributed processing) emerged after the introduction of simplified neurons by McCulloch and Pitts (1943).The basic processing elements of neural networks are called artificial neurons [7], or simply neurons or nodes. In a simplified mathematical model of the neuron, the effects of the synapses are represented by connection weights that modulate the effect of the associated input signals, and the nonlinear characteristic exhibited by neurons is represented by a transfer function [4]. The neuron impulse is then computed as the weighted sum of the input signals, transformed by the transfer function. The learning capability of an artificial neuron is achieved by adjusting the weights in accordance with the chosen learning algorithm. The learning situations in neural networks [8] may be classified into three distinct sorts. These are supervised learning, unsupervised learning, and reinforcement learning. The English sentence recognition system is constructed around the modular architecture of preprocessing, feature extraction and recognition [9]. Mainly the prototype of the system focuses on two concepts first of all testing , a model is built from the sentence drawn by a mouse and recognizing the sentence enforce the Back-Propagation algorithm and after that then displaying the recognized English sentence character. Figure 1 shown the English sentence recognition system.

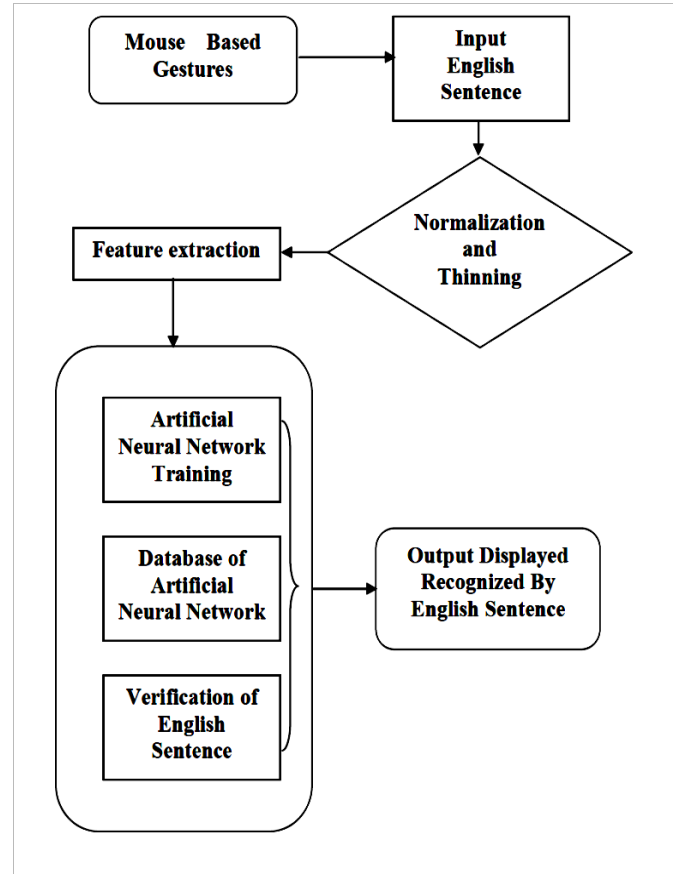

**Fig 1: The Mouse-Based English Sentence Recognition System**

## 3.1 The Pre-Processing

The preprocessing phase normally includes many techniques applied for binarization, thinning (skeletonization), noise removal, skew detection, slant correction, normalization, and contour making like processes to make character image easy to extract relevant features and efficient recognition. These techniques include segmentation to isolate individual characters, thinning (skeletonization), contour marking, normalization, thinning, filtration etc. Which types of pre-processing techniques will suite; it highly depends on our requirements and is also influenced by mechanism adopted in later steps. However many [10] other basic techniques are commonly applied to all applications, for example, noise removal, because that makes image ease to process and recognize in further steps.

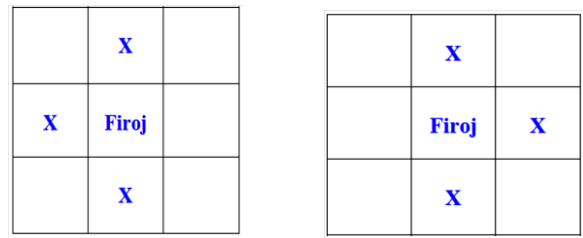

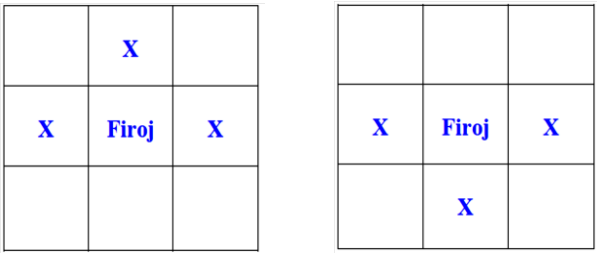

**Fig 2: The framework of an Element in Thinning Algorithm**

Thinning is a morphological operation that is used to remove selected foreground pixels from binary images, somewhat like erosion or opening. It can be used for several applications, but is particularly useful for skeletonization. In this mode it is commonly used to tidy up the output of edge detectors by reducing all lines to single pixel thickness. Thinning is normally only applied to binary images, and produces another binary image as output. The skeleton obtained must have the following properties: must be as thin as possible, connected and centered. When these properties are satisfied, the algorithm must stop. In this paper the Pre-processing phase [11] is comprised of Normalization and Thinning. Before the sentences were presented to artificial neural networks for recognition, they were normalized and thinned and then their feature vector was extracted. The sentence was normalized so that the size of all characters is standardized in order to match the extracted patterns. The thinning algorithm used transforms a picture to a set of simple digital arcs, which false intractable along the measurement axis. It is mainly a search and removing process that removes only those border pixels whose deletion if first case does not change connectivity of their vicinal spot, and second case does not change the length of already thinned capture picture. The thinning algorithm first segregates the border points for each pass. In an alone pass removing of border points are done from four sides in a North, East, South and West sides respectively. If the vicinal of the considered border pixel i.e. "Firoj", match to any of the framework elements shown in figure 2, it is removed and this process goes on until no change takes place on all boundaries.

## 3.2 The Feature Extraction

Feature extraction is used to extract relevant features for recognition of characters based on these features. First features are computed and extracted and then most relevant features are selected to construct feature vector which is used eventually for recognition. The computation of features is based on structural, statistical, directional, moment, transformation like approaches. Feature extraction is extracting information from raw data which is most relevant for classification purpose and that minimizes the variations within a class and maximizes the variations between classes [12]. Selection of a feature extraction method is probably the single most important factor in achieving high recognition performance in character recognition systems. Different feature extraction methods are designed for different representations of the characters, such as solid binary characters, character contours, thinned (skeletons sentence) or gray-level sub images of each individual character. On feature extraction stage each character is represented by a feature vector, which becomes its identity [13]. The major goal of feature extraction is to extract a set of features, which maximizes the recognition rate with the least amount of confusion.

The capture picture is normalized and thinned, it's distinct features are extracted using the Pixel delivery method. The pixel segmentation method is statistical in nature and uses the block method. The centroid of the script is calculated on the basis of pixel delivery. The entire image is divided into 12 blocks of 30° (i.e. Orientation, θ n) each with the centroid as the center of the "circle". In each block, based on the pixel segmentation, the pixel segmentation percentage is calculated with floating-point precision. The slope is calculated for each block first. And then each pixel position is coinciding relative to the slope it is assigned to belong to that distinct block.

$$\theta_n = \tan\left(\frac{A_{n+1} - A_n}{B_{n+1} - B_n}\right)$$

Where, n= 1,2,3,……..,N-1 and N represents the length of the mouse based gesture.

## 4. THE PROPOSD APPROCH FOR GENRAL FRAMWORK ENGLISH SENTENCE RECOGINATION SYSTEM

There are many tradeoffs in designing a handwriting recognition system. At one extreme, the designer puts no constraints on the user and attempts to recognize the user's normal writing [14]. This paper proposes an English sentence recognition system for mouse-based written English sentence in real-time based on Artificial Neural Network. The input of the system is the images of the English sentence through a taken by the mouse-based gestures. In this paper, the neural network technique is used to do the task. The network has been trained with a set of true English sentence. If the feature points are close enough to the trained specimen then the script is regarded as a correct one otherwise an incorrect one. The proposed multilayer feed-forward network structure contains three layers firstly the input layer, secondly hidden layer and a thirdly output layer as shown in figure 3.

The input layer neurons are fully connected [15] to the hidden neurons which in turn are fully connected to the output neurons. The input layer has ANi = 12 neurons corresponding to 12 feature points of each sentence. The hidden layer has ANh = 200 neurons (ANh was determined through experiment) and the output layer has No = 26 neurons corresponding to the different English sentence. Thus, the proposed neural network has total ANi + ANh + No = 12 + 200 + 26 = 238 neurons . Though the network was trained for 4000 epochs, it actually took 10000 iterations for its training. The learning rate and the momentum were set to 0.10 for the network.

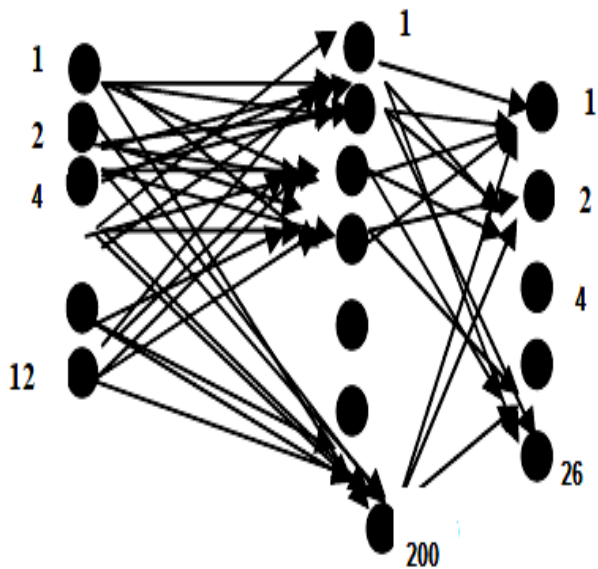

**Fig 3: The Inter connections between neurons of various layers with Artificial Neural Network Structure**

In this paper we are proposing a neural network serving as network classifier, can be using a two types of process the first Back-Propagation training process and the second verification process has done [16]. The Artificial neural network was trained to recognize the English sentence by using the following procedure. The network is produced with training specimens, which consists of a pattern of activities of the input units together with the desired pattern of activities for the output units. After that arbitrate how closely the right output of the network matches the required output. The change weight of each connection so that the network presents a good approximation of the required output. After that we are discovering a set of weights that will enable a given network to compute a given function is usually a nontrivial process. An analytical solution exists only in the normal case of pattern association i.e. When the network is linear and the goal is to map a set of orthogonal input vectors to output vectors and the weights are given by below

$$W_{hlay} = \Sigma \ (iv \ hlay^{pi} \ tv \ hlay^{pi}) / (\| \ iv^{pi} \ \|)$$

Where, iv is an input vector tv is the target vector and pi is the pattern index.

The Backpropagation algorithm is used to learn the weights of a multilayer neural network with a fixed architecture. It performs gradient descent to try to minimize the sum squared error between the network's output values and the given target values [17]. The train a neural network to perform some task, we are must adjust the weights of each monad in such a way that the error between the required output and the actual output is decreased. This process requires that the neural network compute the error derivative of the weights. After that it must calculate how the error changes as each weight is increased or decreased slightly. The backpropagation algorithm is the most widely used method for determining the error derivative of the weights.

The back-propagation algorithm consists of many steps first of all it computes how fast the error transformation as the activity of an output unit are transformed. This error derivative is the difference between the real and the required activity. Secondly it Compute how fast the error transformation as the total input received by an output monad is changed. Thirdly it computes how fast the error changes as a weight on the connection into an output monad [18] are changed. This quantity error derivative of the weights is the answer from step secondly multiplied by the activity level of the monad from which the connection emanates. Lastly it Compute how fast the error changes as the activity of a monad in the previous layer is transformed. This crucial step allows back-propagation to be applied to multilayer networks [19]. When the activity of a monad in the previous layer changes, it affects the activities of all the output monad to which it is connected. So to compute the overall effect on the error, we are adding together all these separate impacts on output monads. But each impact is simple to infer. It is the answer in step secondly multiplied by the weight on the connection to that output monad.

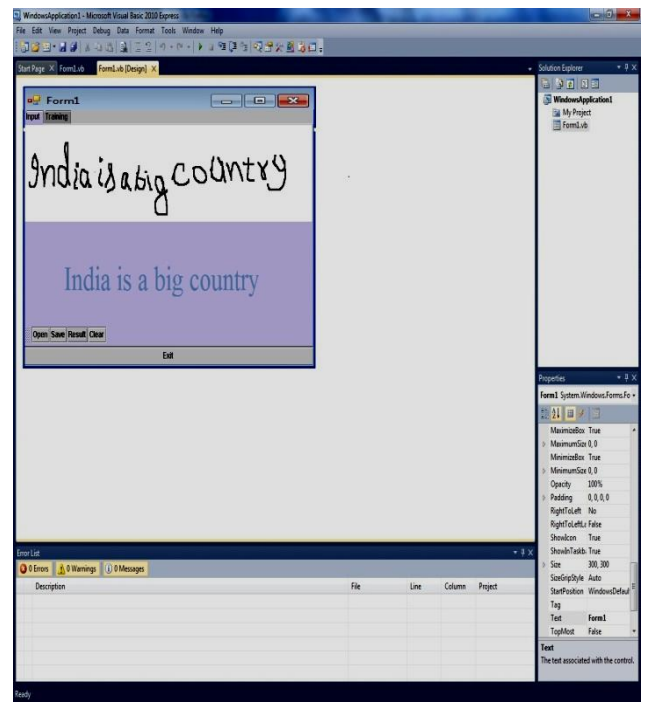

**Fig 4: The Using Artificial Neural Network Screen with Input and Through Mouse-Based Gestures Output**

Finally secondly and lastly steps, we can convert the error derivatives of one layer of units into error derivatives for the former layer [20]. This procedure can be repeated until to get the error derivatives for as many former layers as required. After we know the error derivative of a monad, we can use these steps secondly and thirdly to compute the error derivative of the weights of its incoming connections.

If the feature points are close enough to the trained specimen then the sentence is regarded as a correct one otherwise an incorrect one. The snapshot in figure 4 explains the verification for the mouse drawn English sentence in the English sentence recognition system. In this figure the Input area is first horizontal half of the below given Screen-shot and the input is drawn using the mouse. After that output area in this figure second horizontal half the recognized Sentence is displayed by finally pressing the result button.

## 5. EXPRIMENTAL RESULTES

The experimental results have proven to show excellent recognition rate for both discrete and continuous mouse drawn English sentence. The English sentence recognition system achieves an average recognition rate of 94.1% for isolated English sentence i.e. The sample data collected from 10 persons for 5 English sentences each as given in table 1. The manner in which the graphical representations for table 1 is shown in figure 5 and 6 in five different people as well as similar figure 7 in four different people. The results were generated with Microsoft Visual Basic 2000. The results environment has been established under Windows XP operating system. The hardware included Intel Pentium core 2 duo processor 2.40 GHz, with 2 GB DDR2 RAM, and 320 GB Hard Disk.

**Table 1: The Using Artificial Neural Network Recognition Rate of English Sentence**

| English Sentence | India is a big country |
|---|---|
| Recognition Rate (%) | 94.8%  (2,5,7) |
| English Sentence | Where heritage can be a great unifier as well as divider |
| Recognition Rate (%) | 93%  (1,8) |
| English Sentence | Sport which were traditionally considered a hobby |
| Recognition Rate (%) | 95.2%  (10,3,4) |
| English Sentence | At the school level, sport and fitness are being taken seriously |
| Recognition Rate (%) | 94.3%  (6,9,1) |
| English Sentence | Sports as a field of study is underdeveloped in India |
| Recognition Rate (%) | 93.2%  (5,3,1) |

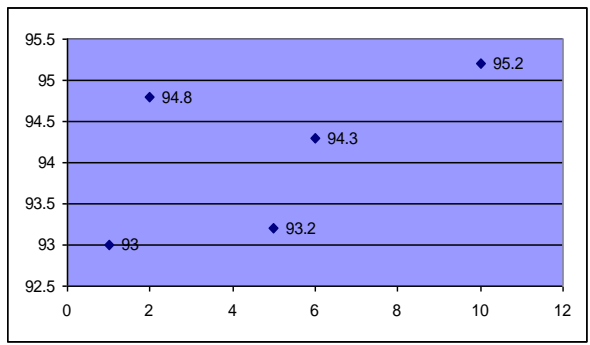

**Fig 5: The Recognition Rate of English Sentence and five different people**

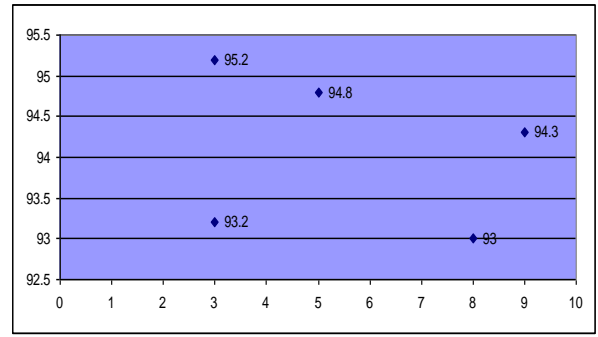

**Fig 6: The Recognition Rate of English Sentence and five different people**

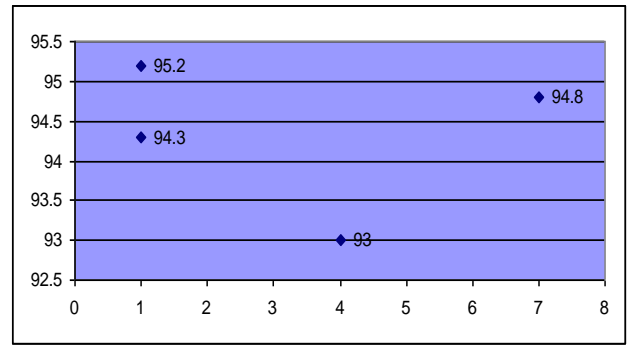

**Fig 7 : The Recognition Rate of English Sentence and four different people**

## 6. CONCLUSION

The Handwriting recognition has been a popular area of research for a few decades under the purview of pattern recognition and image processing. English is a West Germanic language that was first spoken in England and is now the most widely used languages in the world. It is spoken as a first language by a majority of the inhabitants of several nations, including the UK, USA a number of Caribbean nations and other country. Approximately 400 million people speak English as their first language. English today is probably the third largest language by number of native speakers, after Mandarin Chinese and Spanish. It is widely learned as a second language and is an official language of the European Union, many Commonwealth countries and the United Nations, as well as many world organizations. Finally, in this paper we are proposing an English sentence recognition system using Artificial Neural Network through Mouse-Based Gestures, is implemented in Microsoft Visual Basic 2000 and is capable of providing recognition of mouse drawn English sentence in all their forms. The system presents better results for discrete sentence than the continuous one. It should be noted that some of the errors are due to the style of drawing the English sentence mouse and are difficult to be avoided even by a human. Even more, the run time behavior allows the use in real time dynamic applications with a simple personal computer. Though, Back Propagation Algorithm suffers from the drawback of taking too much time for training the neural net still it makes the system provide the most accurate results and hence, more aptly suitable for the kind of multilayer feed forward neural network used for the English sentence recognition system.